%% file: main.tex
\documentclass[sigconf]{acmart}  

\usepackage{multirow}
\usepackage{float}
\usepackage{array}
\usepackage{makecell}
\usepackage{placeins}
\usepackage{lipsum}
\usepackage{balance}

\AtBeginDocument{%
  \providecommand\BibTeX{{%
    \normalfont B\kern-0.5em{\scshape i\kern-0.25em b}\kern-0.8em\TeX}}}

\copyrightyear{2022} 
\acmYear{2022} 
\setcopyright{acmcopyright}\acmConference[MM '22]{Proceedings of the 30th ACM International Conference on Multimedia}{October 10--14, 2022}{Lisboa, Portugal}
\acmBooktitle{Proceedings of the 30th ACM International Conference on Multimedia (MM '22), October 10--14, 2022, Lisboa, Portugal}
\acmPrice{15.00}
\acmDOI{10.1145/3503161.3548277}
\acmISBN{978-1-4503-9203-7/22/10}

\begin{document}

\title[Interactive Video Corpus Moment Retrieval using Reinforcement Learning]{Interactive Video Corpus Moment Retrieval \\ using Reinforcement Learning}

\acmSubmissionID{2265}

\author{Zhixin Ma}
\email{zxma.2020@phdcs.smu.edu.sg}
\affiliation{
  \institution{Singapore Management University}
  \country{Singapore}
}

\author{Chong Wah Ngo}
\email{cwngo@smu.edu.sg}
\affiliation{
  \institution{Singapore Management University}
  \country{Singapore}
}



\begin{abstract}
Known-item video search is effective with human-in-the-loop to interactively investigate the search result and refine the initial query. Nevertheless, when the first few pages of results are swamped with visually similar items, or the search target is hidden deep in the ranked list, finding the know-item target usually requires a long duration of browsing and result inspection. This paper tackles the problem by reinforcement learning, aiming to reach a search target within a few rounds of interaction by long-term learning from user feedbacks. Specifically, the system interactively plans for navigation path based on feedback and recommends a potential target that maximizes the long-term reward for user comment. We conduct experiments for the challenging task of video corpus moment retrieval (VCMR) to localize moments from a large video corpus. The experimental results on TVR and DiDeMo datasets verify that our proposed work is effective in retrieving the moments that are hidden deep inside the ranked lists of CONQUER and HERO, which are the state-of-the-art auto-search engines for VCMR. 
\end{abstract}



\maketitle

\input{introduction}
\input{related_work}
\input{problem_definition}
\input{methodology}
\input{experiment}

\input{conclusion}

\bibliographystyle{ACM-Reference-Format}
\balance

\input{reference}

\input{appendix}

\end{document}

%% file: introduction.tex
\section{Introduction}

``Finding needle from a haystack'' is always the grand challenge of multimedia search. Past research efforts~\cite{thomee2012interactive,lee2021ivist,pevska2021w2vv++,amato2021visione} have demonstrated that interactive search, which allows user to inspect search result and modify query, can significantly boost the performance. With human-in-the-loop, an incomprehensive query can be ``remedied'' in various ways after having hindsight from imperfect search result. For example, relevant keywords overlooked in the initial query can be added; out-of-vocabulary query terms which are interpreted incorrectly by a search engine can be replaced with synonyms; and contextually relevant terms discovered through result inspection can be appended to the original query. Nevertheless, such procedure is always tedious and often requires query modification in a trial-and-error manner. Furthermore, as query history is not traced over time, each modification is treated as an independent search session~\cite{rossetto2019deep,wu2021sql,heller2021towards}, ending up in a lack of a persistent view between the search results of adjacent sessions. Due to these factors, user experience is negatively impacted as reported in the live competition of Video Browser Showdown (VBS)~\cite{lokovc2021reign,rossetto2020interactive}. Consequently, a search target hidden deep in a long video or ranked list can be easily overlooked due to mental tiredness.

This paper addresses the issue of interactive search by simultaneous planning of search navigation path and continuous updating of query based on user feedbacks. Specifically, we imagine that search process is a random walk over the items retrieved by a search engine. The planning of a navigation path seeks to find a route to reach a search target within a few rounds of interaction between user and system. At every interaction, user is recommended an item along the route for comment. Taking the user feedback, the system interactively updates query history and adjusts navigation path, with the long-term goal of shortening the route from the current position to a search target. Intuitively, the planning phase relieves a user from painstaking browsing of search results. The updating of query history captures subtle changes in user feedbacks over time, preventing the drifting of search results over different query modifications.

Path navigation is a classic AI planning problem often implemented with reinforcement learning (RL)~\cite{silver2016mastering,mnih2013playing}. Typically, RL requires an agent to probe the future by exploring different paths to accumulate rewards or experiences for long-term planning. Applying RL for interactive video search, as adopted in this paper, is a non-trivial problem due to the following difficulties. First, the navigation space of a search result is extremely huge even with only a few hundreds of retrieved items to explore. Acquiring sufficient training samples by traversing different paths to reach a search target while collecting user feedbacks along the path is practically intractable. This issue is particularly complex for video search as the query types are open-ended. In addition, there are diverse expressions of search intention for a query and providing feedbacks for a recommended video. Constructing a training set that can tackle the large variations due to various factors (query, user, navigation) is highly challenging. Second, user interaction will last for few iterations only. In general, assuming long period provision of user feedbacks is impractical and can frustrate search experience. This inevitably imposes the constraint that an RL agent should reach the target within limited time steps, rather than steering user traversing a long way for future reward.

To the best of our knowledge, there are limited research efforts exploring RL for interactive search in large video corpus. The existing works apply RL either for single video moment localization~\cite{he2019read,wu2020tree} or domain-specific dialog-based image retrieval~\cite{Guo2018dialogue}. In this paper, we take the first attempt demonstrating that RL using a user simulator has high potential enabling retrieval of search targets hidden deep in a ranked list. The proposed user simulator automatically prompts feedback to a recommended target, providing either a missing concept or pinpointing irrelevant content for training as RL probes the future for navigation planning. We experiment with RL-based interactive search on the challenging task of video corpus moment retrieval (VCMR)~\cite{lei2020tvr,li2020hero,hou2021conquer,zhang2021hammer}. Experimental results show that the proposed user simulator can retrieve the search targets of some hard queries, which are ranked outside the search depth of 100 by VCMR engines such as HERO~\cite{li2020hero} and CONQUER~\cite{hou2021conquer}. When replacing the simulator with human, more hard queries can be resolved within few steps of interaction.
 
The main contribution of this paper is exploring of reinforcement learning for interactive search on large-scale video corpus. Particularly, we demonstrate the first time that, riding on the advances in multimedia content analysis such as concept detection~\cite{wu2020tnterpretable,ueki2018waseda,jiang2009representations} and cross-modal embedding~\cite{lu2019vilbert,tan2019lxmert,li2020hero}, it is realistic to develop user simulation for intelligent multimedia applications to take advantage of human-computer interaction. Using VCMR as a showcase, this paper sheds light on the feasibility of planning a navigation path for interactive search, which is yet to be explored by other systems in venues such as VBS~\cite{lokovc2021reign} and ad-hoc video search~\cite{2021trecvidawad}.

%% file: related_work.tex
\section{Related Work}

Interactive video search has a long trace of history since VideOlympics~\cite{snoek2008videolympics} to joint force human and machine intelligence in multimedia search. Successful attempts include the engagement of users in the search loop to provide relevancy feedback~\cite{yan2003multimedia,kratochvil2020som,lokovc2022video,kratochvil2020somhunter} for the positive examples of search result. Harnessing on the feedbacks, the system interactively trains a classifier on-the-fly to improve search performance. Such paradigm, nevertheless, is only valid for ad-hoc video search~\cite{2021trecvidawad,nguyen2021interactive,snoek2008videolympics}, with assumption of having abundant positive examples to be harvested for training. For known-item search (KIS)~\cite{lokovc2019interactive,lokovc2019framework,zhang2015interactive} or video moment search~\cite{lei2020tvr,li2020hero,hou2021conquer}, where typically only one search target per query, leveraging relevancy feedback to harvest positive examples for query refinement becomes inapplicable. Note that relevancy feedback can also be applied to glean examples visually similar to but not positive of a search target~\cite{kratochvil2020somhunter,cox1996pichunter}. While these examples are effective for refinement of search space, they are not suitable for query refinement. As user has a concrete information need and has encountered the search target in a different occasion, the query tends to be verbose, which poses challenge to query understanding. As lesson learnt from the past few years of VBS benchmarking activities~\cite{lokovc2021reign,nguyen2021interactive}, providing user interface for efficient search result navigation is essential. Particularly, an appealing interface can make user stay in focus to interact with search results and reformulate query~\cite{rossetto2020interactive,zhang2015interactive,lokovc2019viret}. However, the search efficacy drops with increasing of data size, especially when the targets are outside of a search range that can hardly be reached by user.

Reinforcement learning (RL) has been actively researched recently for video moment localization~\cite{ma2021hierarchical,cao2020strong} and temporal grounding of natural language in videos~\cite{he2019read,wu2020tree}. Nevertheless, these efforts are devoted to localization of a moment or clip-of-interest from a single untrimmed video. Specifically, given a video and a query expressed in natural language, RL is exploited to iteratively move a sliding window forward or backward along the time axis until reaching a video segment that best matches the query semantics. Basically, an agent takes an action and receives a reward or penalty at every iteration before planning the next action. The existing works vary in terms of learning environment, size and types of action space. For instance, \cite{wu2020bar} formulates RL for weakly supervised learning, where the training examples consist of only video-query pairs without the labelling of temporal boundaries. In addition to temporal movement (e.g., moving left or right), some works consider dynamic expansion and shrinking of a sliding window~\cite{he2019read}, speed of temporal movement~\cite{wu2020tree}, scale of spatial object and scene~\cite{cao2020strong} as the action space. Tree-structured policy learning is also investigated in \cite{wu2020tree} for progressive decision of actions in a coarse-to-fine manner.

Our proposed work is largely different from the existing works~\cite{he2019read,wu2020tree,ma2021hierarchical,wu2020bar,cao2020strong} as we address the issue of retrieving a moment-of-interest from a large corpus containing a few thousands of untrimmed videos. The size of action space is not restricted to temporal movement within a single video, but also include ``jumping” across videos to locate a search target. More importantly, previous works do not consider user feedback and the environment of RL is relatively static. Specifically, the policy network learning is mainly driven by the ground-truth locations of moments~\cite{he2019read,wu2020tree}, without considering the dynamics of environment where user may response differently to an action. In this paper, the action space is built upon a graph constructed over the videos in a corpus. The action space is time-varying rather than static as in~\cite{he2019read, wu2020tree, cao2020strong}, depending on the navigated paths over time on the graph. In addition, the system state is also determined by user feedbacks. In other words, a state will change dynamically not only based on the action taken but also the feedback prompted by user on the action.

As inclusion of user feedbacks in RL generally demands large training samples, the advantage of feedbacks is only explored in the domain-specific dataset for dialog-based product search~\cite{Guo2018dialogue,Yu2019VisionLanguageRV}. Capitalizing on large dialog conversation in the dataset for training, \cite{Guo2018dialogue,Yu2019VisionLanguageRV} demonstrate the retrieval of products by modeling of user feedbacks over multiple dialog turns. In \cite{Guo2018dialogue}, relative captioning, which contrasts the visual discriminativeness between two images with a caption, is explored as user feedbacks by RL for interactive retrieval of product images. Nevertheless, the construction of a relative captioning dataset as in \cite{Guo2018dialogue} for videos is highly difficult. Different from product image search, the video content is much richer and diverse. Without query context, relative captioning that only narrates one aspect of the visual difference between two video moments cannot accurately capture the user search intention. Instead of recruiting crowdsourcing workers to provide captions as feedbacks~\cite{das2017visual,Guo2018dialogue}, this paper proposes user simulation to provide concepts as feedbacks for training. As a target moment is, by definition, a “known item” that has been previously watched by the searcher, the simulator basically imitates the searcher to comment missing or irrelevant concepts in a recommended moment. Comparing to relative captioning~\cite{vedantam2017context,Guo2018dialogue}, the requirement of understanding query context to provide feedback is also relaxed by use of concepts.

Compared to single-video moment localization, video corpus moment retrieval (VCMR) is a more challenging task. However, the existing works~\cite{li2020hero,hou2021conquer,lei2020tvr,zhang2021hammer,escorcia2019temporal} considers VCMR an auto-retrieval problem without human in the loop. The performance is hardly satisfied where around 60\% of the target moments in TVR~\cite{lei2020tvr} and DiDeMo~\cite{hendricks2017didemo} datasets, are ranked outside the search depth of 100 by state-of-the-art search engines~\cite{li2020hero,hou2021conquer}. In this paper, we study the extent in which RL-based interactive search can retrieve these search targets hidden deep inside a rank list.

%% file: problem_definition.tex
\section{Problem Statement}
\label{sec:problem_statement}
A user issues a natural language query and is provided with a ranked list of candidate moments retrieved by a search engine. The navigation of target moment starts when the user selects a moment for browsing. Denote the user query as $q_0$ and the selected moment as $m_0$. The goal is to plan for a navigation path that reaches the target moment $m^*$ from $m_0$ within a limited number of time steps. Specifically, along the path, the system recommends a moment for browsing while the user provides keyword-based feedback specifying either a missing or irrelevant concept in the moment. Based on the feedback, the navigation path is adjusted dynamically in every step and the system takes an action that ideally shortens the time step required to reach $m^*$.

To this end, we define the problem of interactive moment navigation as a Markov Decision Process (MDP) with the tuple $(\mathcal{S}, \mathcal{A}, \mathcal{R}, \mathcal{P})$. $\mathcal{S}$ is the set of states, and $\mathcal{A}$ is the set of actions that adjusts the navigation path by selecting a new moment and enters a new state. $\mathcal{R}$ is a reward function to merit an action that successfully shortens a navigation path. $\mathcal{P}$ is the state transition probability, where the system maintains a graph $\mathcal{G}$ outlining the navigation space between any two moments. The problem is to seek an optimal path from $m_0$ to $m^*$ over $\mathcal{G}$ based on user feedbacks and gain the largest accumulated reward. In the following section, we will detail the policy network for action selection, learning of reward function and construction of graph $\mathcal{G}$ for navigation.

%% file: methodology.tex
\input{Figs/fig_tex_architecture}

\section{Interactive Moment Retrieval}
Figure~\ref{fig:architecture} depicts the framework of interactive search, which is composed of agent and environment. The agent performs query update based on user feedback and interactively refines the navigation path by selecting an action (i.e., moment) over a graph $\mathcal{G}$. Based on the recommended moment, the environment evaluates the search progress, provides feedback, and generates a reward for the agent.

\subsection{Navigation Over Graph}
As user has a search goal in mind, the navigation of moments within or across videos is not random but based on their content relatedness. We build a graph that models the multi-modal similarity of moments over a large video collection for navigation planning. A graph $\mathcal{G}=(\mathcal{V}, \mathcal{E})$ outlines all the possible paths of traversal between any two moments on $\mathcal{G}$. The distance between the initial moment $m_0$ and the target moment $m^*$, denoted as $d$, is defined by the number of traversed edges along the path from $m_0$ to $m^*$. The navigation strategy is to plan for the shortest path, corresponding to the optimal value of $d$ or $d^*$, such that a user can navigate to the target moment with the least number of steps. Note that $\mathcal{G}$ is an undirected graph. Planning is challenging as the size of candidate paths between $m_0$ and $m^*$ is expected to be large. 

The graph $\mathcal{G}$ is constructed by first connecting temporally adjacent clips in a video and then extending to clips within and across videos depending on their semantic similarity measured with transformer features \cite{li2020hero} and concept features \cite{wu2020tnterpretable}. 
Considering that the clips within a video tend to be more similar than those across videos, we maintain an empirical ratio such that, for every node in $\mathcal{G}$, its incident edges should span across clips in different videos (see section~\ref{subsec:dataset_and_setting} for details). Note that the definition of moment is query dependent, and ideally a graph should be constructed during query time to capture per-query moment candidates and their pairwise navigation paths. However, dynamic construction of per-query graph is computationally expensive and practically infeasible. Instead, we build a static graph capturing the navigation paths of every two clips in the dataset. During query time, the mapping between moments and the nodes on $\mathcal{G}$ is established based on their video identities and time stamps.

\subsection{Reinforcement Learning}
\label{subsec:rl}
The agent is essentially a policy network $\pi_{\theta}$ that predicts the probabilistic distribution of actions based on the status of system and the graph $\mathcal{G}$. Recall that interactive search is modeled as a MDP with the tuple $(\mathcal{S}, \mathcal{A}, \mathcal{R}, \mathcal{P})$. Denote $s_t \in \mathcal{S}$ as a system state at time $t$, defined as following
\begin{equation*}
    s_t = (q_0, m_t, {F}_t), \,\, s_0 = (q_0, m_0)
\end{equation*}
where $q_0$ is the initial query, ${F}_t=\{f_i\}_{i=1}^t$ captures the feedback history up to time $t$, and $m_t$ denotes the currently visited moment. At $t=0$, $s_0$ comprises of only $q_0$ and a moment $m_0$ picked by a user from the search result for inspection. 

Given the current state $s_t$ and the neighboring nodes of $m_t$ on $\mathcal{G}$, the network $\pi_\theta$ determines the next action $a_t \in \mathcal{A}_t$ by picking $m_{t+1}$ within the observation window of $s_t$ (Section~\ref{subsubsec:policy_net}). During training time, as the ground-truth target moment is known, a reward function is proposed to evaluate $m_{t+1}$ (Section~\ref{subsubsec:reward_func}). Meanwhile, the environment (or user) will assess $m_{t+1}$ by providing feedback indicating the deviation of $m_{t+1}$ from $m^*$. With these, as shown in Figure~\ref{fig:architecture}, the agent updates the query and enters the next state $s_{t+1}$ for refinement of navigation path.

\subsubsection{Policy Network}
\label{subsubsec:policy_net}
In principle, an agent can recommend any moment to user at time $t$. In other words, the action space $\mathcal{A}_t$ can be as large as the number of retrieved moments for a query. Nevertheless, such strategy is computationally impractical for requiring the reranking of all candidates and will hamper real-time delivery of search results. Instead, we employ graph $\mathcal{G}$ to narrow the action space for $s_t$. The space is defined by an observation window that includes the moments that are reachable by $m_t$ within a distance that requires $k$ number of edge traversals. Denote the set of moments as $\mathcal{N}_{m_t}^{k}$, the action space comprises the neighbors of $m_t$ reachable within $k$ traversals. Note that the cardinality $|\mathcal{N}_{m_t}^{k}|$ increases exponentially with the value of $k$, implying a larger search space at each timestamp but potentially fewer steps to reach from $m_t$ to $m^*$ over time. 

The policy network is implemented as a cross-modal neural network that projects the updated query $q_t$ and a candidate moment $m_i \in \mathcal{N}_m^{k}$ into a joint space for similarity measure. Let $u_i$ as the similarity between $q_t$ and $m_i$, the policy function $\pi_\theta$ is as following
\begin{equation}
\label{eq:policy_net_1}
    u_i = \, <FC_m(\mathbf{m_i}) \cdot FC_q(\mathbf{q})>, \, m_i \in \mathcal{A}_t
\end{equation}
\begin{equation}
\label{eq:policy_net_2}
    \pi_\theta(s_t) = [u_0, ..., u_{|\mathcal{A}_t|}]
\end{equation}
where $<\cdot>$ denotes cosine similarity. The projections of query and moment are implemented with two multi-layer feedforward networks, $FC_m$ and $FC_q$, respectively. To keep a balance between exploitation and exploration~\cite{sutton2018reinforcement}, $\pi_\theta$ will randomly select a $m_{t+1}$ with a probability of $\epsilon$ and the moment with the highest similarity score with $1-\epsilon$ probability.

\subsubsection{Reward Function}
\label{subsubsec:reward_func}
The goal of agent is to reach $m^*$ as quickly as possible by traversing the edges on $\mathcal{G}$. Hence, the reward function is designed to merit agent whenever an action $a_t$ shortens the traversed distance between $m_t$ and $m^*$. The degree of merit depends on the actual distance from $m_{t+1}$ and $m^*$ as well as the number of iterations taken so far up to time $t$. Ideally, the action $a_t$ should reach the target moment if $m^*$ is within the observation window of $s_t$. Otherwise, the distance from $m_{t+1}$ to $m^*$ should be shortened after taking the action $a_t$. Denote $d_t$ as the shortest distance from $m_t$ to $m^*$ on $\mathcal{G}$, the action $a_t$ is evaluated by the reward function as following
\begin{equation}
\label{eq:reward}
r_t =
    \begin{cases}
      \frac{1}{2^{d_{t+1}}}-\phi\cdot t, & d_t > d_{t+1}\\
      -\phi\cdot t, & d_t = d_{t+1}\\
      -\frac{1}{2}-\phi\cdot t, & d_t < d_{t+1}
    \end{cases} 
\end{equation}
where the agent receives a positive reward if $d_{t+1} - d_{t} < 0$ and otherwise. The amount of reward is controlled by $d_{t+1}$ and a penalty term $\phi$ which deducts reward by a constant factor magnified by the number of time steps. Intuitively, an agent will receive higher reward when moving closer to $m^*$ with a smaller number of steps. 

The progressive traversal of $\mathcal{G}$ to recommend moments can be viewed as a sequential decision-making problem with the long-term goal of moving user closer to the target. In interactive search, however, a user will be frustrated if the target moment cannot be reached just within few rounds of interaction~\cite{tan2019drill}. Hence, even if the long-term planning with the help of user feedbacks can eventually reach the target, the user experience will be negatively impacted if taking more than acceptable number of steps. To this end, we define long-term reward with a discounting factor $\gamma$ as following
\begin{equation}
\label{eq:big_reward}
R_t =
    \begin{cases}
      r_t+\gamma\cdot Q_w(s_t, a_t), & t=T_{max} \\
      r_t+\gamma\cdot R_{t+1}, & t = 0, ..., T_{max}-1
    \end{cases} 
\end{equation}
where the agent iterates at most $T_{max}$ rounds or until reaching $m^*$. The $Q_w(s_t)$ value function is learnt to predict the reward when the agent iterates for $T_{max}$ rounds. $Q_w$ is implemented as a multi-layer feedforward neural network as
\begin{equation}
    Q_w(s_t) = FC_w([FC_m(\mathbf{m_t}); FC_q(\mathbf{q})])
\end{equation}
where $[;]$ denotes the concatenation of the projected query and video features. The training of $Q_w$ will be further discussed in Section~\ref{subsubsec:policy_net_learning}.

\subsection{Implementation}
\label{subsec:implementation}

\subsubsection{User Simulator}
\label{subsubsec:user_simulator}
Collecting feedbacks for training agent under reinforcement learning setting is challenging in real-world. Employing user simulator has become a common practice for auto generation of training samples~\cite{li2016user, schatzmann2006survey}. We propose a user simulator to automatically prompt a keyword highlighting the prominent visual difference between a recommended moment $m_{t+1}$ and the target moment $m^*$. The keyword indicates either a concept present in $m^*$ but missing in $m_{t+1}$ or vice versa, i.e., an irrelevant concept present in $m_{t+1}$. 

The simulation is enabled by having each moment being indexed with the top-50 most relevant semantic concepts and their probability distribution~\cite{wu2020tnterpretable}. When evaluating the action $a_t$, the user simulator investigates $m_{t+1}$ and picks a concept, either from $m_{t+1}$ or $m^*$, that deviates most in terms of probability values. The sign of deviation indicates whether a selected concept is suggested to be included or removed from the next recommendation of moment. To introduce feasibility, we adopt $\epsilon$-greedy strategy that, with a probability of $\epsilon$, the simulator will select a concept either missing in $m_{t+1}$ or not present in $m^*$ in random as feedback.

\subsubsection{Query Update}
\label{subsubsec:query_update}
The initial query $q_0$ is represented as a vector, $\boldsymbol{q_0}= [BERT(q_0);GRU(q_0)]$, concatenated by the features extracted from BERT~\cite{devlin2018bert} and GRU~\cite{chung2014gru}, respectively. The feedback $f_t$, which is composed of a keyword, is represented as a one-hot vector with sign indicating whether the corresponding concept should be expanded to or ignored from the current query $q_t$.  To keep track of the history of user feedbacks, we adopt a GRU cell for query update as following
\begin{equation}
    \boldsymbol{q}_{t+1} = GRU(\boldsymbol{f_t}, \, \boldsymbol{q_t}) + \boldsymbol{q_0}\, , \ t > 0
\end{equation}
\begin{equation}
    \boldsymbol{f}_{t} = \boldsymbol{W_{1}}\cdot \mathbb{I}_{f_t}+\boldsymbol{b}_{1}
\end{equation}
where $\mathbb{I}_{f_t}$ is a one-hot feedback vector, $\mathbf{W}_1$ and $\mathbf{b}_1$ are learnable parameters.

\subsubsection{Policy Network Learning}
\label{subsubsec:policy_net_learning}

We adopt on-policy learning to train the agent by sampling trajectories (or navigation paths) from $m_0$ to $m^*$ using Monte Carlo Tree search (MCTS)~\cite{sutton2018reinforcement}. A sampling session terminates when the agent finds the ground-truth $m^*$ or traverses for $T_{max}$ iterations. We adopt both supervised learning and model-based policy learning for reinforcement learning. For the former, the agent is trained with triplet loss objective with $m^*$ as the positive example and the remaining moments in the corresponding observation window as negative samples, as following
\begin{equation}
\label{eq:triplet_loss}
    L_{triplet} = max(0, c + u^+ - u^-)
\end{equation}
\begin{equation}
    u^{\{+/-\}} = <FC_m(\mathbf{m}^{\{+/-\}})\cdot FC_q(\mathbf{q})>
\end{equation}
where $m^+$ and $m^-$ denote positive and negative samples respectively. The parameter $c$ is the margin and $<\cdot>$ is the similarity function in Equation~\ref{eq:policy_net_1}. Note that supervised learning considers only the samples where $m^*$ falls within the observation window of $m_0$ for training. To enable training on the entire trajectories of samples from $m_0$ to $m_*$, we employ the advantage actor-critic algorithm (A2C)~\cite{Mnih2016asynchronous}.

Using the advantage function, the expected advantage $J_a$ is maximized as following
\begin{equation}
    J_a = \sum_{a \in \mathcal{A}}\pi_\theta(a|s)(R-Q_w(s))
\end{equation}
\begin{equation}
    \nabla_{\theta} J_a \approx \sum_{t} \nabla_{\theta} log\,\pi_\theta(a_t|s_t) (R_t-Q_w(s_t))
\end{equation}
Using stochastic gradient descent (SGD), the optimization problem is equivalent to the minimization of $L_{policy}$ loss, as following
\begin{equation}
    L_{policy} = -\sum_{t} log\,\pi_\theta(a_t|s_t) (R_t-Q_w(s_t))
\end{equation}
Note that the value function $Q_w$ is also trained to predict reward at every time step with the following loss function
\begin{equation}
    L_{mse} = \sum_{t}(Q_w(s_t)-R_t)^2
\end{equation}
 to minimize the mean square error between the predicted and actual rewards. Finally, the three loss functions are linearly weighted as the overall loss, i.e.,
\begin{equation}
\label{eq:loss_func}
    Loss = \lambda_1\, L_{triplet} + \lambda_2\, L_{policy} + \lambda_3\, L_{mse}
\end{equation}

%% file: Figs/fig_tex_architecture.tex
\begin{figure*}[!htbp]
    \center
    \begin{minipage}{2. \columnwidth}
        \centerline{\includegraphics[width=0.8 \columnwidth]{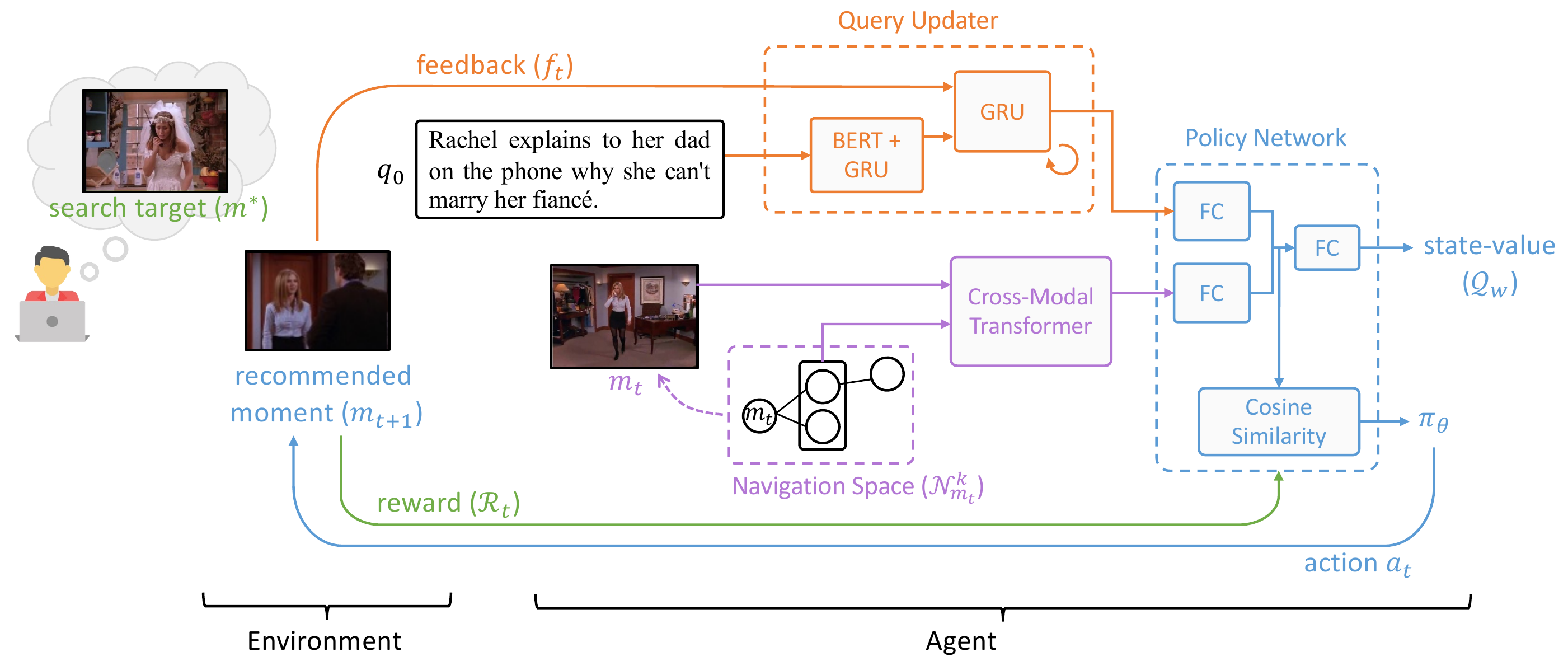}}
    \end{minipage}
    \caption{Proposed architecture: the agent recommends a video moment ($m_{t+1}$) that best fits the system state ($q_0$, $f_t$, $m_t$) based on the navigation space at time $t$. The environment (user simulator) evaluates the recommendation and provides feedback ($f_t$) as well as reward ($r_t$) for training policy network.}
    \label{fig:architecture}
\end{figure*}

%% file: experiment.tex
\input{Tabs/tab_sta_dst}

\section{Experiment}
\input{Tabs/tab_main_result}
The proposed work is validated by first comparing to different options of implementation to claim the merits of user simulation and policy learning (Section \ref{subsec:comparison}). The user simulator is verified by further showing its effectiveness in locating search targets that are ranked low by HERO~\cite{li2020hero} and CONQUER~\cite{hou2021conquer} (Section \ref{subsec:rl_vs_manual_browse}). Finally, the performance difference between simulator and human is presented to provide insights beyond using a single keyword as feedback and a moment as recommendation (Section~\ref{subsec:manual_feedback}).

\subsection{Datasets and Settings}
\label{subsec:dataset_and_setting}
The experiments are conducted on two large datasets, TVR~\cite{lei2020tvr} and DiDeMo~\cite{hendricks2017didemo}, following the standard split of training, validation and testing sets as listed in Table~\ref{tab:sta_dst}. TVR consists of professional-edited videos from six TV episodes, while the videos on DiDeMo are unedited and randomly sampled from YFCC100M Flickr videos~\cite{thomee2016yfcc100m}. Note that, as the testing set of TVR is not publicly available, the results are reported on the validation set. 

To plan for navigation path, one graph is constructed for each dataset. On TVR, the videos are segmented into clips according to timestamped subtitles. These clips are linked based on their visual and textual similarities. The former is measured based on the transformer features~\cite{li2020hero} and concept features~\cite{wu2020tnterpretable}. The later is measured based on the TF-IDF scores of subtitles. As the clips within a video tend to have higher similarity, the between-video clips are linked more loosely with a lower threshold of similarity when constructing the graph $\mathcal{G}$. We set an empirical ratio such that, for each node in a graph, the node will have around 60\% of edges connecting to the nodes from other videos. The remaining 40\% of edges will connect to nodes from the same video. As the clips within a video are more similar to each other, the edges of a node could connect to all the nodes from the same video if without this empirical ratio. The empirical ratio is a hyper parameter that is learnt from the training data. To this end, $\mathcal{G}$ consists of 147,985 nodes, with each node having on average 3.98 edges connecting the clips of 3.04 different videos. The memory consumption of both graph index and node feature is around 2 GB and the disk storage is around 6.5 GB during inference. On DiDeMo, we use the 5-second clips provided by the dataset to build the graph. As no subtitles are provided, the segments are linked using only the visual features~\cite{li2020hero,wu2020tnterpretable}. The graph is composed of 114,923 nodes and 263,609 edges. Similarly, each node has 4.59 edges connecting the clips of 2.61 different videos. The memory consumption is around 1.5 GB and the disk storage is around 2.6 GB. Note that a graph will be split into multiple subgraphs for training and testing according to the split of videos in a dataset.

The inputs to policy network are visual features extracted from \cite{li2020hero} and textual features from BERT~\cite{devlin2018bert} and GRU~\cite{chung2014gru}. In addition, we employ the dual-task network in \cite{wu2020tnterpretable} to extract the top-50 concepts for each clip. The user simulator will sample keywords from the concepts as feedbacks, as discussed in Section~\ref{subsubsec:user_simulator}. A total of 277,689 and 129,004 trajectories are extracted from TVR and DiDeMo, respectively, for policy network training. In all the experiments, $T_{max}$ is fixed to $7$ assuming that user and agent will interact at most seven rounds. By default, the observation window of a state $s_t$ is set to $k=3$. This setting corresponds to an average size of action space with 67.3 and 57.8 moments on TVR and DiDeMo, respectively. The parameters of policy network are empirically set as: $\epsilon = 0.1$ for all greedy policies following the convention~\cite{mnih2013playing}, $\theta = 0.01$ (Equation~\ref{eq:reward}), $\gamma = 0.8$ (Equation~\ref{eq:big_reward}), $c= 0.1$ (Equation~\ref{eq:triplet_loss}) and $\lambda_{\{1,2,3\}} = \{1,0.1,0.1\}$ (Equation~\ref{eq:loss_func}). 

\input{Figs/fig_tex_results_step_simulation}

\subsection{Performance Comparison}
\label{subsec:comparison}

\input{Tabs/tab_recall_real_rank}
As interactive search for video corpus moments is a new task without prior work, we compare the proposed work against three baselines: Random, No-feedback and Imitation. All the methods use the same $\mathcal{G}$ for navigation planning. For Random, the agent randomly picks a moment at $t$ based on the observation window of $s_t$ until reaching the target $m^*$. In contrast, No-feedback selects a moment most similar to the query $q_t$ using cosine similarity and the network is trained with Equation~\ref{eq:triplet_loss}. As no feedback mechanism is considered, $q_t=q_0$ throughout the interaction. Imitation is similar to our proposed work, except using the imitation learning strategy~\cite{hussein2017imitation}. Specifically, the training involves only triplet loss in Equation~\ref{eq:triplet_loss}. We use recall@1 as the performance measure since each query has only one ground-truth moment.

In this section, the initial moments, i.e., $m_0$, are sampled from the graph $\mathcal{G}$ for experiments. Specifically, we sample four moments with $d_0$={1,2,3,4} distance from $m^*$, where $d_0$ denotes the minimum number of edge traversal between $m_0$ and $m_*$. The experiments involve a total of 98,070 and 40,766 queries on TVR and DiDeMo datasets, respectively. Table~\ref{tab:main_res} lists the average recall@1 performance (plus standard deviation) for all the queries. Note that recall@1 = 1 if an agent can locate $m^*$ within $T_{max}=7$ interactions, and otherwise recall@1 = 0. As expected, the performance drops as $d_0$ increases. Random performs poorly due to large action space at each step $t$. By leveraging the initial query $q_0$ to recommend the best possible moment over different time steps, No-feedback boosts the performance sharply. Our approach based on advantage actor-critic (A2C) further pushes the performance by 2.03\% of improvement over No-feedback. For queries with $d_0 \leq 3$, more than 45\% of their $m^*$ can be located by our approach. The result drops to around 25\%, nevertheless, when $d_0$ = 4. This is because the observation window is set to $k=3$, and a minimum of two interactions is required to reach $m^*$ when $d_0$ = 4. Consequently, if the first recommendation is deviated from the shortest path of $m_0$ to $m^*$, the chance of reaching $m^*$ within $T_{max}$ interactions become lower. When comparing to Imitation learning, A2C also shows almost consistently higher recall@1 performance for queries across different values of $d_0$. The result basically indicates the advantage of long-term path planning using the trajectories sampled by MCTS for training. The speed of our approach is similar to Imitation. No-Feedback is about 1.7 times faster due to no update of query model. On average, our approach will take 0.007 second per iteration and 0.05 second per query (7 rounds of iterations) with the user simulator on a desktop with a single RTX 3090 GPU.

Figure \ref{fig:res_tvr_step_simulation} further details the number of interactions required to reach the search target $m^*$. As shown, except Random, all the compared approaches can retrieve the search targets for more than 35\% of queries. The recall@1 performance improves with the increase of time steps. The performance gap between different approaches also becomes larger with the increase of time steps. This basically verifies the advantage of having user feedback and adopting A2C for policy network training. Figure~\ref{fig:example_simulation} illustrates how the simulator manages to locate the target of a query. The initial moment $m_0$ shows the character Ryan in an indoor scene, where `squirrel' and `tree' specified in the query are missing. The simulator adds the concept `sword', which is contextually relevant to $m^*$, as feedback and the agent recommends $m_1$ with Ryan holding sword. Although $m_1$ is not the target, it resides in the same video as $m^*$, and more importantly, $m^*$ falls in the observation window of $m_1$. Consequently, the agent finds $m^*$ in the next step when the feedback `leaf' is added.

\input{Figs/fig_tex_example_simulation}


\subsection{RL-based versus manual browsing}
\label{subsec:rl_vs_manual_browse}

\input{Figs/fig_tex_example_browse_didemo}

The objective of this section is to investigate the effectiveness in searching target moments that are hidden deep inside a ranked list. The experiment is conducted by having $m_0$ provided by the HERO~\cite{li2020hero} and CONQUER~\cite{hou2021conquer} search engines. Specifically, $m_0$ refers to the top-1 rank moment retrieved by a search engine. As we are only interested in searching moments that cannot be easily located by the auto-search engine, the experiments involve only those queries whose ground-truth moments are ranked at and outside of depth@10 by these search engines. We cluster these queries into four groups based on the ranks of their targets, as shown in Table~\ref{tab:recall_real_rank}. On TVR, by our approach, 14\% of queries whose targets are between the ranks of 10-50 can be successfully retrieved. As expected, the performance drops gradually for the queries whose targets are ranked behind in the list. Nevertheless, our approach is still able to perform reasonably well by retrieving 8\% of search targets that are ranked beyond 200th position. Similar performance trend is also observed on DiDeMo, but with lower recall rate due to unavailability of subtitles in the dataset for similarity measure. 
Figure~\ref{fig:example_browse_didemo} shows query examples where the simulator is able to locate their targets. In Figure~\ref{fig:example_browse_didemo} (a), the target is hidden in 700th position of the CONQUER ranked list . The top-1 retrieved moment ($m_0$) shows a `box' in the scene but without `creature'. The simulator suggests removing `car' from $m_0$, which leads the agent arrives at $m_1$ that resides in the same video as the target. The moment ($m_1$) contains `box' but not `creature'. After two more rounds of interaction, the agent locates the target. Similarly, for the example in Figure~\ref{fig:example_browse_didemo} (b), the simulator suggests having an outdoor scene by adding `outside' but removing `car' and `street' in the subsequent interactions, before arriving at the target. While the result is encouraging, the simulator cannot distinguish concepts salient to queries, which results in several failure cases. For example, when general concepts such as `guy' and `object' are picked by the simulator, the agent may easily navigate away from the shortest path to $m^*$.

We use the rank-step plot to show the number of steps required to reach $m^*$ at a given rank on TVR. Figure~\ref{fig:res_tvr_scatter_conquer} visualizes the distribution of queries whose search targets are successfully retrieved by our approach on the rank-step plot. As a comparison, the manual browsing effort from $m_0$ to $m^*$ is proportional to the depth of $m^*$ in a ranked list. By our approach, the targets at a depth beyond 100 can be retrieved by the proposed user simulator within seven steps of interaction.

\input{Figs/fig_tex_res_sactter_real}

\subsection{Automatic versus manual feedback}
\label{subsec:manual_feedback}
\input{Figs/fig_tex_example_manual_concept}

In this section, instead of using the user simulator, a human is instructed to interact directly with the system. Note that this is not a user study. The experiment aims to provide insights regarding the practical effectiveness of a user simulator. We sample 50 queries on TVR which cannot be retrieved by the simulator in Section~\ref{subsec:rl_vs_manual_browse} for experiments. The targets of these queries are equally distributed in five intervals of the search range from 20th to beyond 200th position. In the experiment, the human subject is first asked to view the ground-truth moment and then start interacting with the agent by picking the top-1 retrieved moment by CONQUER as $m_0$. Like the simulator, one keyword per moment is suggested by the subject. Among the 50 queries, 18\% of their targets are successfully located by the human, on average using only 3.22 steps. Figure~\ref{fig:example_manual_concept} shows an example contrasting feedbacks provided by the human and user simulator. As the current design of simulator does not consider query context, the simulator can provide feedback contradicting to the initial query (e.g. remove `standing'). When the concepts are not correctly detected by~\cite{wu2020tnterpretable}, the provided feedbacks (e.g. `suit', `brown') drive the agent further away from the target. Human, on the other hand, is sensitive to the prominent difference between the target and recommended moments, by providing feedbacks to remove `phone' and include location `kitchen' and person dressing in `red' to arrive at the target.

To demonstrate the practicality of our approach, we conduct another experiment on these 50 queries by asking the human subject to provide more than one keyword per moment. Surprisingly, the recall rate remains the same despite 4.2 keywords per moment on average are provided. Nevertheless, the average steps of interaction is reduced to 3.08. Furthermore, we also experiment the setting of allowing the agent to recommend five moments in one interaction. In this setting, the human subject can pick one out of the five suggested moments for feedback, including picking one of the top-5 ranked moments by the search engine as $m_0$. The feedback can include more than one keyword. The recall@1 performance is boosted to 44\% and the average steps is further reduced to 2.05. Out of the 22 retrieved targets, the original ranks of 12 targets are beyond 100th position. We notice that, by selecting the most suitable $m_t$ to provide feedback, the result will be positively impacted. Human can also recognize the context of query well and provide the salient concept to supplement a selected $m_t$. This multiplying effect partially addresses the limit that, when the only recommended $m_t$ is weakly relevant, arbitrary feedback could be provided, which eventually misleads the agent to deviate from the optimal path to $m^*$. Most failure cases are due to no partially relevant moments found in the top-5 retrieved moments. In this case, the selection of $m_0$ becomes arbitrary. Further providing feedbacks may not be able to revert the navigation path towards the search target. Note that the time human spending on investigating the recommended moments is not strictly proportional to the number of recommendations. Instead, approximately the same interaction time is spent for some queries due to a shorter navigation path to reach to the target. In the experiment, the human subject spends only 10 seconds more per query for five recommendations than for one recommendation.

%% file: Tabs/tab_sta_dst.tex
\begin{table}[]
\caption{Dataset statistic showing the number of videos and queries in different splits.}
\label{tab:sta_dst}
\centering

\begin{tabular}{lcccccc}
\toprule
Dataset & \multicolumn{3}{c}{\#Video} & \multicolumn{3}{c}{\#Query} \\ \cmidrule(r){2-4} \cmidrule(r){5-7}
        & Train    & Val     & Test    & Train   & Val     & Test    \\ \hline
TVR     & 17,435   & 2,179   & -       & 87,175  & 10,895  & -       \\
DiDeMo  & 8,395    & 1,065   & 1,004   & 32,624  & 4,160   & 3,982   \\ 
\bottomrule
\end{tabular}

\end{table}

%% file: Tabs/tab_main_result.tex
\begin{table*}[!h]
\caption{Model comparison of recall@1 on TVR and DiDeMo dataset.}
\label{tab:main_res}
\centering

\begin{tabular}{lcccccccc}
\toprule
Model            & \multicolumn{4}{c}{TVR}                                                      & \multicolumn{4}{c}{DiDeMo}                \\ \cmidrule(r){2-5} \cmidrule(r){6-9}
            & $d_0$=1 & $d_0$=2 & $d_0$=3 & $d_0$=4         & $d_0$=1 & $d_0$=2 & $d_0$=3 & $d_0$=4 \\ \hline
Random      & 0.058 $\pm$0.001 & 0.044 $\pm$0.001 & 0.024 $\pm$0.001 & 0.007 $\pm$0.001     
            & 0.093 $\pm$0.004 & 0.078 $\pm$0.003 & 0.053 $\pm$0.005 & 0.023 $\pm$0.001 \\
            
No-Feedback & 0.489 $\pm$0.011 & 0.475 $\pm$0.013 & 0.442 $\pm$0.012 & 0.204 $\pm$0.008                
            & 0.259 $\pm$0.004 & 0.242 $\pm$0.001 & 0.229 $\pm$0.005 & 0.121 $\pm$0.002 \\
            
Imitation   & 0.514 $\pm$0.011 & 0.485 $\pm$0.010    & \textbf{0.448} $\pm$0.011 & 0.210 $\pm$0.006     
            & 0.275 $\pm$0.005 & 0.261 $\pm$0.009    & 0.242 $\pm$0.008          & 0.132 $\pm$0.006 \\
            
Ours        & \textbf{0.534} $\pm$0.016 & \textbf{0.495} $\pm$0.002 & 0.446 $\pm$0.005           & \textbf{0.216} $\pm$0.003    
            & \textbf{0.296} $\pm$0.003 & \textbf{0.275} $\pm$0.002 & \textbf{0.262} $\pm$0.009  & \textbf{0.141} $\pm$0.003 \\

\bottomrule
\end{tabular}
\end{table*}




%% file: Figs/fig_tex_results_step_simulation.tex
\begin{figure}[h]
    \center
    \begin{minipage}{0.8 \columnwidth}
        \centerline{\includegraphics[width=1 \columnwidth]{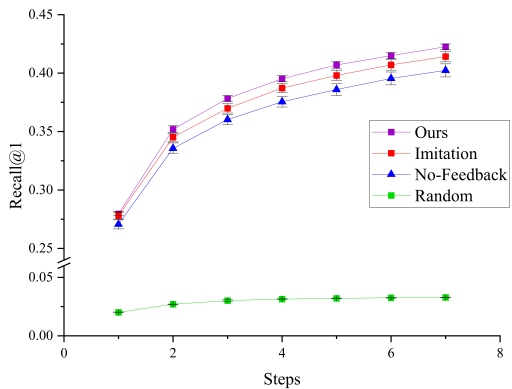}}
    \end{minipage}
    \caption{The step-wise recall@1 on TVR dataset.}
    \label{fig:res_tvr_step_simulation}
\end{figure}

%% file: Tabs/tab_recall_real_rank.tex
\begin{table*}[!htbp]
\caption{Performance of identifying search targets from the ranked lists of HERO and CONQUER. The second row shows the search depths of targets in a ranked list. The last two rows show the distribution of targets across search depths. The number inside parenthesis indicates the percentage of targets being located by our approach.   }
\label{tab:recall_real_rank}
\centering

\begin{tabular}{lwc{1.3cm}wc{1.3cm}wc{1.3cm}wc{1.3cm}wc{1.3cm}wc{1.3cm}wc{1.3cm}wc{1.3cm}}
\toprule
        & \multicolumn{4}{c}{TVR}                                                      & \multicolumn{4}{c}{DiDeMo}                                                \\ \cmidrule(r){2-5} \cmidrule(r){6-9}
        & (10-50]        & (50-100]           & (100-200]           & $>$200           & (10-50]           & (50-100]           & (100-200]           & $>$200      \\ \cmidrule(r){1-9}
HERO    & 2179 (14\%)    & 822 (9\%)          & 785 (6\%)           & 328 (5\%)        & 683 (9\%)         & 379 (5\%)          & 396 (4\%)           & 886 (3\%)   \\
CONQUER & 2049 (14\%)    & 681 (9\%)          & 416 (9\%)           & 98 (8\%)         & 823 (8\%)         & 415 (7\%)          & 388 (3\%)           & 591 (3\%)   \\ \bottomrule
\end{tabular}

\end{table*}

%% file: Figs/fig_tex_example_simulation.tex
\begin{figure}[!h]
    \center
    \begin{minipage}{1. \columnwidth}
        \centerline{\includegraphics[width=1 \columnwidth]{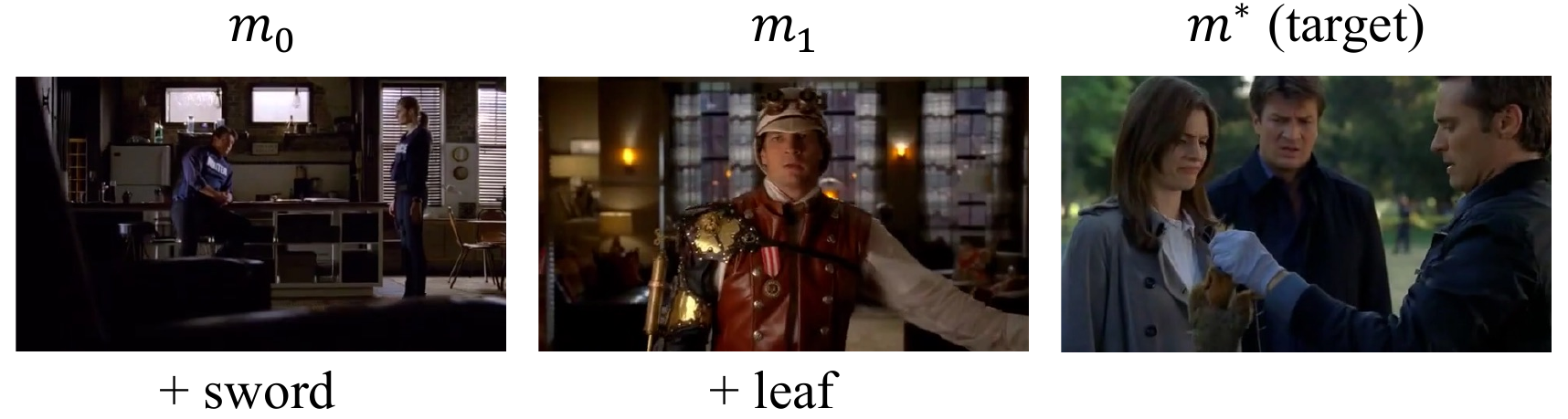}}
    \end{minipage}
    \caption{Example of user simulator interacting with agent to search for the target moment of the query ``Ryan carries a dead squirrel down from a tree and gives it to someone''. The symbom `+' indicates a concept added by the simulator as feedback.}
    \label{fig:example_simulation}
\vspace{-0.3cm}
\end{figure}


%% file: Figs/fig_tex_example_browse_didemo.tex
\begin{figure*}[!htbp]
    \center
    \begin{minipage}{2. \columnwidth}
        \centerline{\includegraphics[width=0.85 \columnwidth]{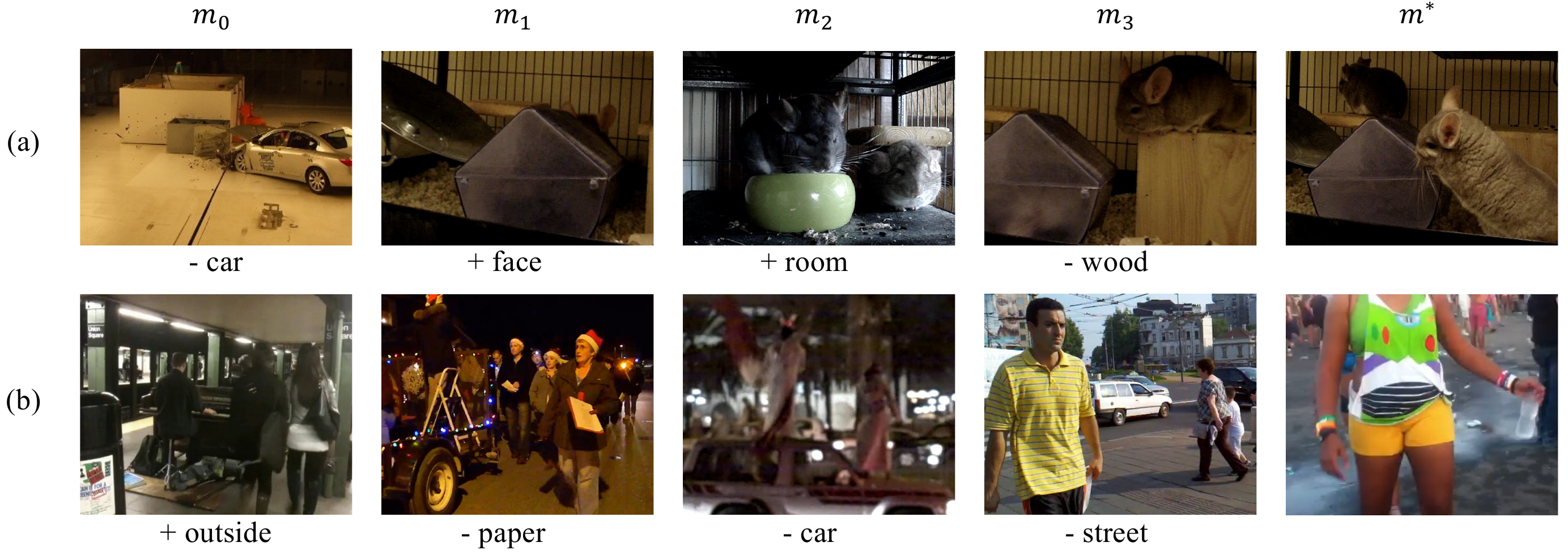}}
    \end{minipage}
    \caption{Examples of the user simulator interacting with agent for queries on DiDeMo: (a) ``the small creature jumps out of the box'', (b) ``girl with green shirt walks by''. The symbols `+' and `-' indicate the concepts being suggested to add or remove, respectively, for a recommended moment.}
    \label{fig:example_browse_didemo}
\end{figure*}

%% file: Figs/fig_tex_res_sactter_real.tex
\begin{figure}[h]
    \center
    \begin{minipage}{.8 \columnwidth}
        \centerline{\includegraphics[width=1 \columnwidth]{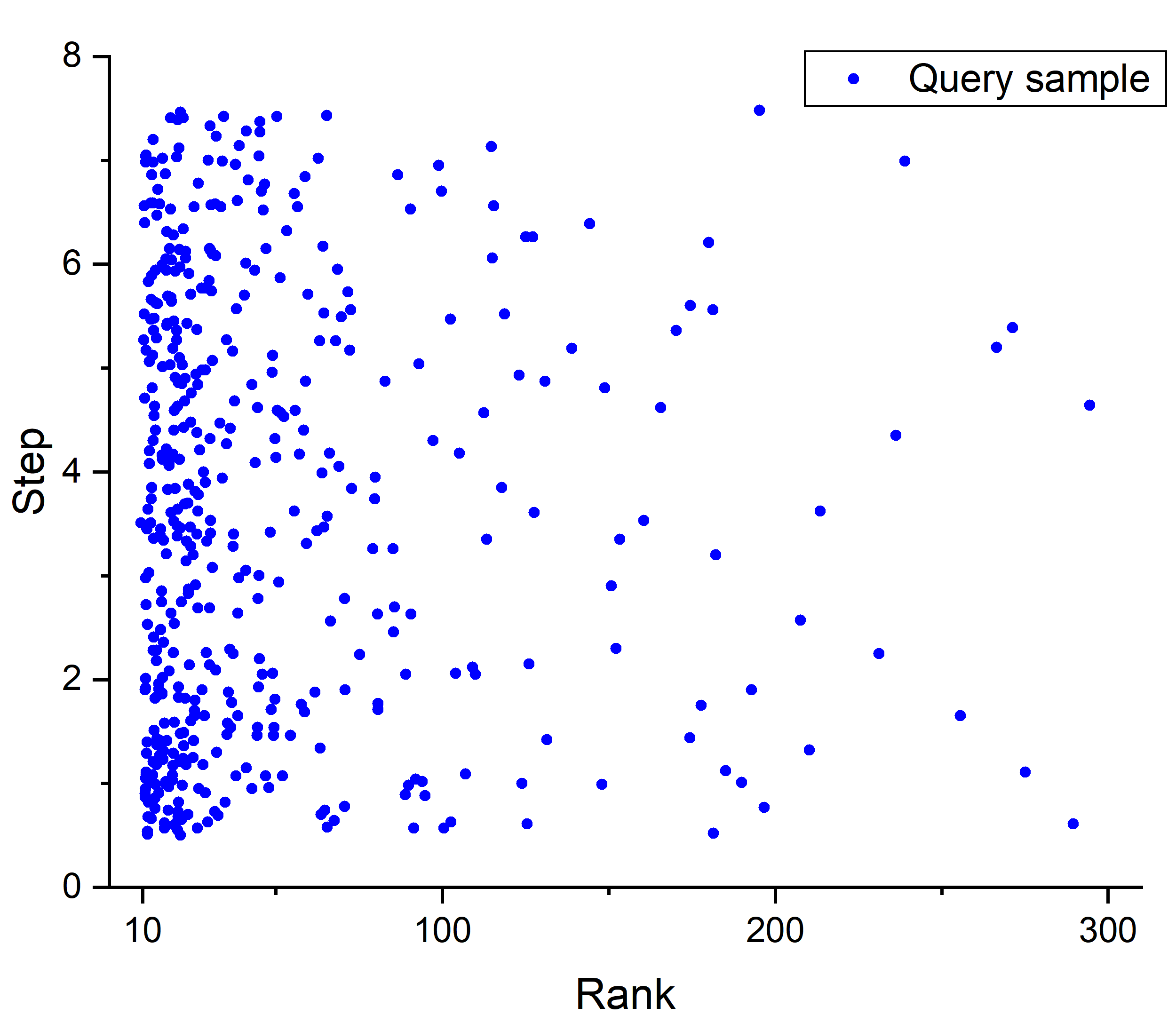}}
    \end{minipage}
    \caption{Rank-step plot showing the number of steps required to locate a target from a search depth ranked by CONQUER. Each point (X, Y) shows a query sample whose target is ranked at X position and retrieved with Y steps.}
    \label{fig:res_tvr_scatter_conquer}
\end{figure}

%% file: Figs/fig_tex_example_manual_concept.tex
\begin{figure}[h]
    \center
    \begin{minipage}{1. \columnwidth}
        \centerline{\includegraphics[width=.9 \columnwidth]{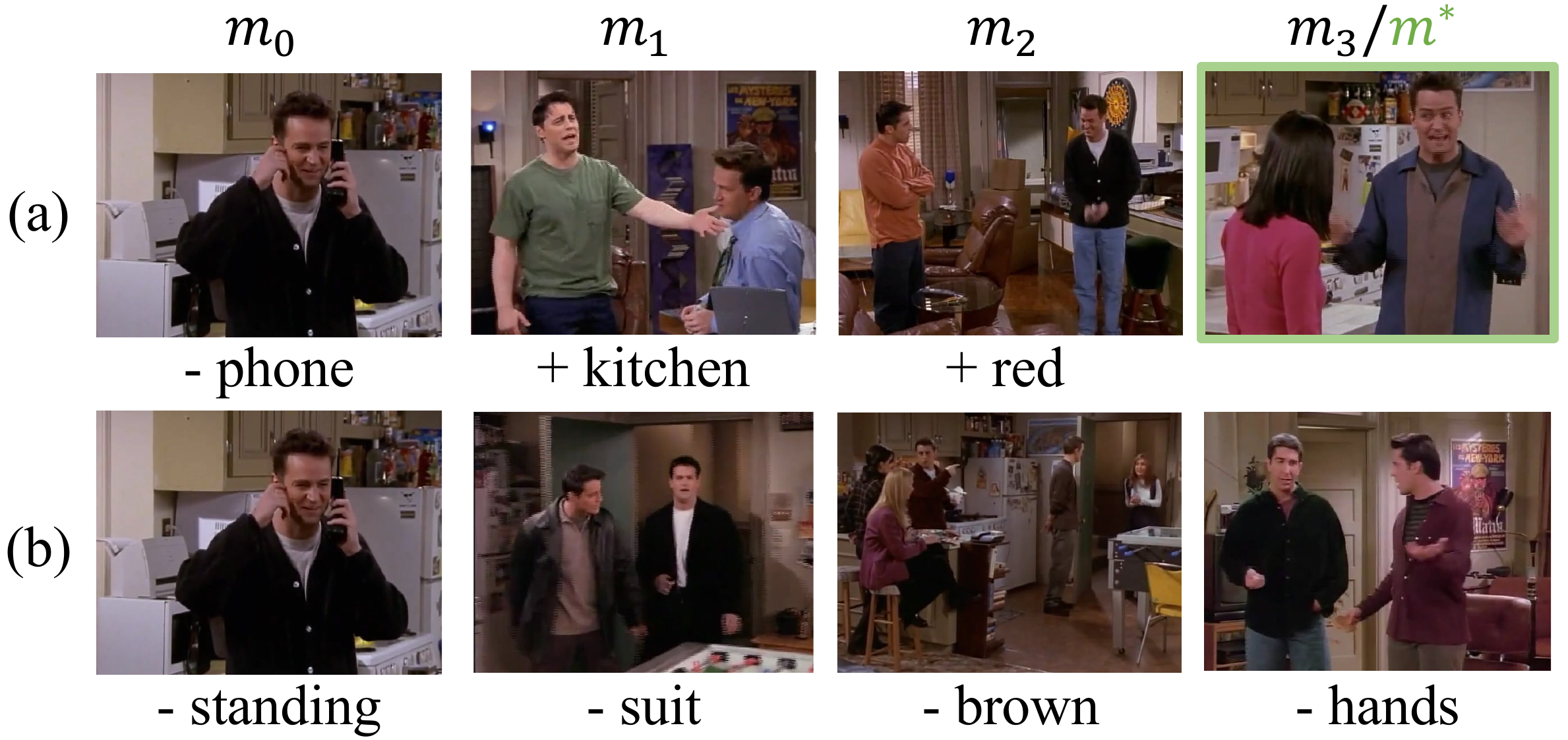}}
    \end{minipage}
    \caption{Example comparing how (a) human and (b) user simulator interact with the agent for the query ``Chandler shakes his hands in the air while standing in the kitchen''. The symbols `+' and `-' indicate the concepts being suggested to add or remove for a recommended moment. The search target is marked with green border.}
    \label{fig:example_manual_concept}
\end{figure} 


%% file: conclusion.tex
\section{Conclusion}

Interactively finding search targets not effectively retrieved by automatic search is a tedious task. Our work provides empirical evidence showing the feasibility of exploiting reinforcement learning to speed up the retrieval of targets that are ranked low by search engines. Particularly, the proposed framework with a user simulator interacting with an agent enables planning and navigation of search space without using additional labels for training. As shown in the experiments, the user simulator can be flexibly replaced by human to select a moment for feedback during inference time. Our findings show the substantial difference between human and simulator in terms of performance and search behaviour. Future extensions include devising the simulator to more vividly mimic human behaviour, such as selection of query-aware and salient concepts as feedback, to optimize policy network training. The current work cannot be directly extended for large datasets such as V3C~\cite{luca2019v3c} without training data. Further research is required to adapt the user simulator to different datasets.

\section*{Acknowledgment}
This research was supported by the Singapore Ministry of Education (MOE) Academic Research Fund (AcRF) Tier 1 grant.

%% file: reference.tex

%% file: appendix.tex
\appendix

\input{Tabs/tab_sup_ablation}

\input{Figs/fig_tex_sup_action_sapce_tvr}



\section*{Ablation Study}
We conduct studies to examine the impacts of different parameters on the performance.

\section{Penalty term and discount factor}
For further analysis, Table~\ref{tab:ablation_study} provides comparison of different $\phi$ in Equation~\ref{eq:reward} and $\gamma$ in Equation~\ref{eq:big_reward} in A2C algorithm, respectively. To test the model robustness to the hyper-parameters, we fix one parameter and change the value of the other to see the performance variation. The value of $\phi$ varies from $0.05$ to $0.2$ and $\gamma$ is from $0.7$ to $0.9$. Overall, our model shows its robustness to different values of hyper-parameters. Specifically, on the TVR dataset, the fluctuation of recall@1 performance is less than 2\%. Similarly, the recall@1 performance on the DiDeMo dataset is also stable against the variations in hyper-parameters.


\section{Action space}
Figure~\ref{fig:sup_action_sapce_tvr} and~\ref{fig:sup_action_sapce_didemo} show how the size of action space ($\mathcal{N}^k$) impacts retrieval performance. By varying the observation window $k$ from 3 to 6, the average size of action space increases from 67.3 to 2535.6 on TVR and from 57.8 to 1144.7 on the DiDeMo dataset. In general, a smaller action size facilitates retrieval of a search target near $m_0$. Larger action size, on the other hand, is helpful when a search target is further away from $m_0$. This is simply because less number of steps (or interactions) is required to move from $m_0$ to $m^*$ with larger action space. However, this also increases the difficulty of recommendation due to the large number of moment candidates to select. In practice, the sensitivity of action size can be alleviated by recommending multiple moments, each from a different size of action space, for user to select and provide feedback.


\input{Figs/fig_tex_sup_action_space_didemo}

%% file: Tabs/tab_sup_ablation.tex
\begin{table*}[!h]
\caption{Ablation study on the hyper-parameters $\phi$ and $\gamma$.}
\label{tab:ablation_study}
\centering

\begin{tabular}{wc{.8cm}wc{.8cm}wc{1.2cm}wc{1.2cm}wc{1.2cm}wc{1.2cm}wc{1.2cm}wc{1.2cm}wc{1.2cm}wc{1.2cm}}
\toprule
 & & \multicolumn{4}{c}{TVR}                                               & \multicolumn{4}{c}{DiDeMo}                                           \\ \cmidrule(r){3-6} \cmidrule(r){7-10}
 $\phi$ & $\gamma$ & $d_0$=1         & $d_0$=2         & $d_0$=3         & $d_0$=4         & $d_0$=1         & $d_0$=2         & $d_0$=3        & $d_0$=4         \\ \hline
0.1    & 0.8   & \textbf{0.5113} & \textbf{0.4922} & \textbf{0.4527} & \textbf{0.2122} & 0.2959 & 0.2746 & \textbf{0.2620} & \textbf{0.1411} \\ \hline
0.2    & \multirow{2}{*}{0.8} & 0.4991          & 0.4790           & 0.4416          & 0.2073          & 0.2932          & 0.2709          & 0.2536         & 0.1306 \\
0.05 & & 0.5019          & 0.4829          & 0.4430           & 0.2094          & 0.2753          & 0.2626          & 0.2390          & 0.1284 \\ \hline
\multicolumn{1}{c}{\multirow{2}{*}{0.1}} & 0.9  & 0.4990           & 0.4793          & 0.4404          & 0.2020           & 0.2856          & 0.2671          & 0.2490          & 0.1336 \\
\multicolumn{1}{c}{}    & 0.7  & 0.5030           & 0.4794          & 0.4451          & 0.2116          & \textbf{0.2989} & \textbf{0.2858}          & 0.2600           & 0.1410 \\
\bottomrule
\end{tabular}

\end{table*}

%% file: Figs/fig_tex_sup_action_sapce_tvr.tex
\begin{figure*}[!h]
    \center
    \begin{minipage}{2. \columnwidth}
        \centerline{\includegraphics[width=1.0 \columnwidth]{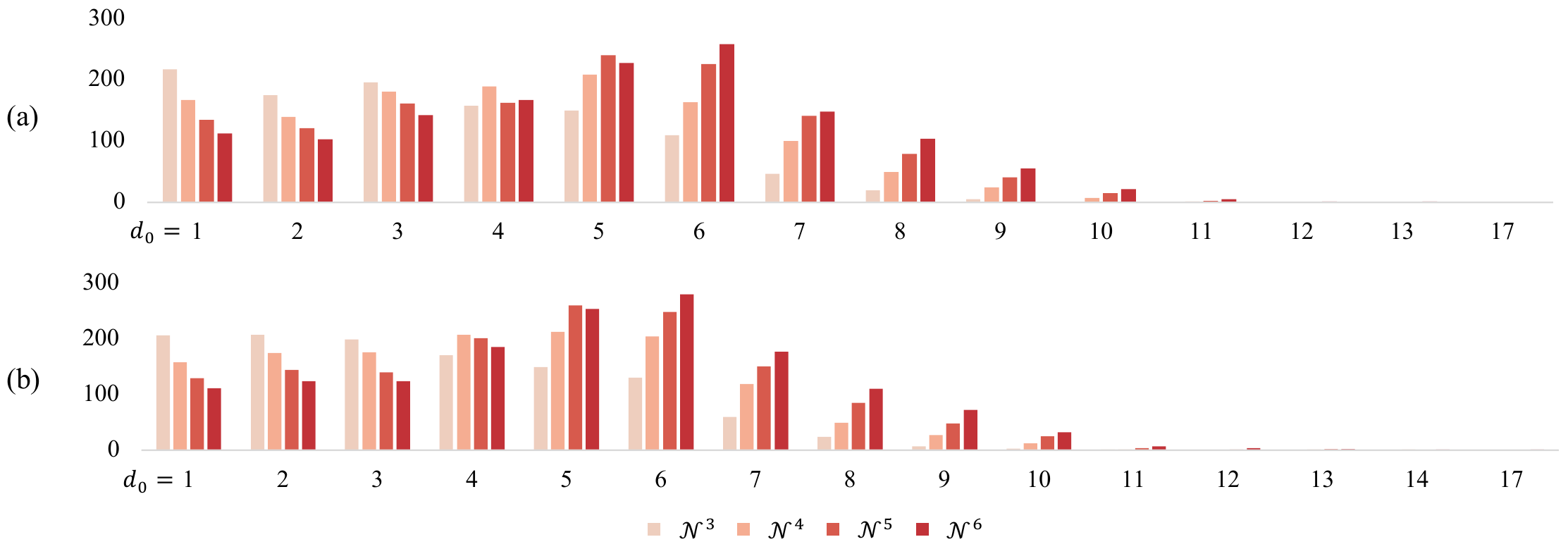}}
    \end{minipage}
    \caption{Retrieval performance by varying the size of action space ($\mathcal{N}^3$ to $\mathcal{N}^6$) for retrieving the search targets at different distances ($d_0=1,...,17$) on TVR dataset based on (a) CONQUER and (b) HERO ranked lists. The y-axis shows the number of queries where their search targets are retrieved by our approach.}
    \label{fig:sup_action_sapce_tvr}
\end{figure*}


%% file: Figs/fig_tex_sup_action_space_didemo.tex
\begin{figure*}[!h]
    \center
    \begin{minipage}{2. \columnwidth}
        \centerline{\includegraphics[width=1.0 \columnwidth]{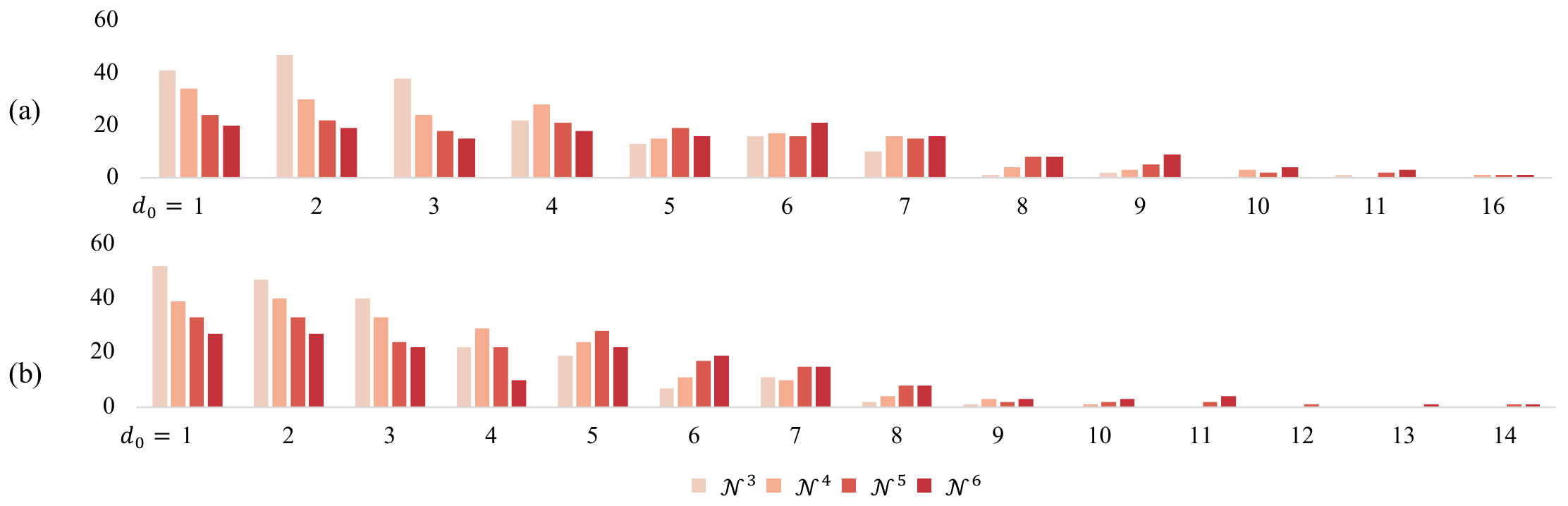}}
    \end{minipage}
    \caption{Retrieval performance by varying the size of action space ($\mathcal{N}^3$ to $\mathcal{N}^6$) for retrieving the search targets at different distances ($d_0=1,...,16$) on DiDeMo dataset based on (a) CONQUER and (b) HERO ranked lists. The y-axis shows the number of queries where their search targets are retrieved by our approach.}
    \label{fig:sup_action_sapce_didemo}
\end{figure*}